\PassOptionsToPackage{unicode}{hyperref}
\PassOptionsToPackage{hyphens}{url}

\documentclass[11pt]{article}

\usepackage[margin=1in]{geometry}

\usepackage{xcolor}
\usepackage{amsmath,amssymb}

\usepackage{iftex}
\ifPDFTeX
  \usepackage[T1]{fontenc}
  \usepackage[utf8]{inputenc}
  \usepackage{textcomp} 
\else 
  \usepackage{unicode-math} 
  \defaultfontfeatures{Scale=MatchLowercase}
  \defaultfontfeatures[\rmfamily]{Ligatures=TeX,Scale=1}
\fi

\usepackage[authoryear]{natbib}
\usepackage{lmodern}
\ifPDFTeX\else
\fi
\IfFileExists{upquote.sty}{\usepackage{upquote}}{}
\IfFileExists{microtype.sty}{
  \usepackage[]{microtype}
  \UseMicrotypeSet[protrusion]{basicmath} 
}{}
\makeatletter
\@ifundefined{KOMAClassName}{
  \IfFileExists{parskip.sty}{%
    \usepackage{parskip}
  }{
    \setlength{\parindent}{0pt}
    \setlength{\parskip}{6pt plus 2pt minus 1pt}}
}{
  \KOMAoptions{parskip=half}}
\makeatother
\usepackage{longtable,booktabs,array}
\usepackage{calc} 
\usepackage{etoolbox}
\makeatletter
\patchcmd\longtable{\par}{\if@noskipsec\mbox{}\fi\par}{}{}
\makeatother
\IfFileExists{footnotehyper.sty}{\usepackage{footnotehyper}}{\usepackage{footnote}}
\makesavenoteenv{longtable}
\usepackage{graphicx}
\makeatletter
\newsavebox\pandoc@box
\newcommand*\pandocbounded[1]{
  \sbox\pandoc@box{#1}%
  \Gscale@div\@tempa{\textheight}{\dimexpr\ht\pandoc@box+\dp\pandoc@box\relax}%
  \Gscale@div\@tempb{\linewidth}{\wd\pandoc@box}%
  \ifdim\@tempb\p@<\@tempa\p@\let\@tempa\@tempb\fi
  \ifdim\@tempa\p@<\p@\scalebox{\@tempa}{\usebox\pandoc@box}%
  \else\usebox{\pandoc@box}%
  \fi%
}
\def\fps@figure{htbp}
\makeatother
\usepackage{bookmark}
\IfFileExists{xurl.sty}{\usepackage{xurl}}{} 
\hypersetup{
  hidelinks,
  pdfcreator={LaTeX via pandoc}}

\author{
	Roland Baatz$^{1,2}$ \\[4pt]
	\parbox{0.9\textwidth}{
		\small
		$^{1}$Leibniz Centre for Agricultural Landscape Research (ZALF),  
		Eberswalder Str. 84, 15374 Müncheberg, Germany.\\
		$^{2}$Current address: CDM Smith, Bouchestr. 12, 12435 Berlin, Germany.\\[4pt]
		\texttt{arxivbaatz@posteo.de}
	}
	}

\date{} 

\title{Generalization and Feature Attribution in Machine Learning Models for Crop Yield and Anomaly Prediction in Germany}

\begin{document}
\maketitle

\begin{abstract}This study examines the generalization performance and interpretability of machine learning (ML) models used for predicting crop yield and yield anomalies in Germany’s NUTS-3 regions. Using a high-quality, long-term dataset, the study systematically compares the evaluation and temporal validation behavior of ensemble tree-based models (XGBoost, Random Forest) and deep learning approaches (LSTM, TCN).

While all models perform well on spatially split, conventional test sets, their performance degrades substantially on temporally independent validation years, revealing persistent limitations in generalization. Notably, models with strong test-set accuracy, but weak temporal validation performance can still produce seemingly credible SHAP feature importance values. This exposes a critical vulnerability in post hoc explainability methods: interpretability may appear reliable even when the underlying model fails to generalize.

These findings underscore the need for validation-aware interpretation of ML predictions in agricultural and environmental systems. Feature importance should not be accepted at face value unless models are explicitly shown to generalize to unseen temporal and spatial conditions. The study advocates for domain-aware validation, hybrid modeling strategies, and more rigorous scrutiny of explainability methods in data-driven agriculture. Ultimately, this work addresses a growing challenge in environmental data science: how can we evaluate generalization robustly enough to trust model explanations?

\end{abstract}

\section{Introduction}\label{introduction}

Prediction in agricultural systems presents both technical and
scientific challenges. Two dominant modeling paradigms have emerged:
process-based models, which rely on mechanistic understanding of
biophysical processes, and machine learning (ML) approaches, which
extract patterns directly from data \citep{lischeid_machine_2022, 	lobell_climate_2011, cao_mapping_2025, tamayo-vera_advanced_2025}. A third,
emerging direction involves integrating these paradigms by leveraging
the interpretability and domain knowledge embedded in process-based
models with the flexibility and predictive power of ML \citep{lischeid_machine_2022,tamayo-vera_advanced_2025,  reichstein_deep_2019}. Recent work in theory-guided data science
highlights the potential of such hybrid models to improve
generalization, particularly in complex environmental and agricultural
systems \citep{reichstein_deep_2019, karpatne_theory-guided_2017}. All modeling approaches ultimately aim to
generalize well, producing reliable outputs under novel conditions by
extrapolating from calibrated or learned parameters. Regardless of
approach, the overarching goal is to build models that generalize well
producing reliable outputs under novel or unobserved conditions. In
process-based models, parameters are typically fixed and tuned through
calibration and assimilation techniques \citep{baatz_reanalysis_2021, 	baatz_evaluation_2017, vrugt_treatment_2008}. In ML
models, numerous parameters are optimized through hyperparameter tuning
\citep{zoller_benchmark_2019, letoffe_shap_2024}. A persistent and critical question remains: how
can we determine whether a model generalizes well enough for scientific
or operational use?

Yield and anomaly prediction in agriculture remains an active area of
research, driven by the increasing availability of high-resolution
spatiotemporal data \citep{cao_mapping_2025, jabed_crop_2024}. These data offer the
potential to improve predictive accuracy across diverse agro-climatic
regions and growing seasons. Traditionally, crop models have been used
to simulate yield by explicitly modeling physiological and environmental
interactions \citep{webber_no_2020}. However, their practical application
is often constrained by over-parameterization and limited use of
observational data in multi-criteria calibration, which in practice can
shift them toward probabilistic behavior rather than strictly
deterministic process representation \citep{seidel_towards_2018}.

In parallel, machine learning has gained significant traction due to its
ability to model complex, nonlinear relationships in data without
requiring prior assumptions about system dynamics \citep{jabed_crop_2024}.
A growing number of studies now use ML to predict yield and yield
anomalies, identify key predictors, and quantify the relative importance
of features \citep{cao_mapping_2025, brill_exploring_2024}. However, a persistent challenge
across many of these studies is the lack of robust, independent model
evaluation particularly evaluations that test generalization across time
or under extreme conditions. This limitation raises important concerns
about the reliability of ML-based feature interpretations.

To answer these questions, the input data and preprocessing steps are
outlined, hereafter the configuration of machine learning models, and
the validation strategy, including both internal and temporally
independent assessments. Then model performance is evaluated across
spatial and temporal dimensions and the reliability of the resulting
feature importance interpretations is assessed. In doing so, this study
aims to support more transparent, generalizable, and interpretable
applications of machine learning in agricultural modeling.

\section{Methodology}\label{methodology}

\subsection{Study region}\label{study-region}

The study focuses on Germany, using its administrative NUTS-3 regions as
the spatial unit of analysis. These regions are well documented in
agricultural statistics and have been widely used in previous studies,
making them a valuable testbed for both crop models and machine learning
applications \citep{lischeid_machine_2022, webber_no_2020, brill_exploring_2024, volker_long-term_2022}.

\subsection{Data}\label{data}

As data source for yield, the study uses published yield data collected
from the German statistical office and harmonized by \citet{volker_long-term_2022}. Winter wheat yield data comes in kg per hectares
at annual resolution from 1979 to 2022 for NUTS-3 region where
available. Meteorological data was provided by the German Weather
Service at 1km grid size and daily resolution for mean, minimum and
maximum air temperature, precipitation, wind speed, solar radiation and
relative humidity. The product interpolates station measurements at 1km
resolution and daily aggregates. A winter wheat crop mask was used for
the year 2018 only based on a composite product of IACS, remote sensing
and machine learning \citep{blickensdorfer_mapping_2022}. Based on the masked data by
NUTS-3 region, NUTS-3 region specific weather indicators were
calculated. These weather indicators yielded weekly, monthly and
quarterly averaged for each region as in Table 1. In addition extreme
weather indicators were calculated following the methodology outlined in
the Thünen report \citep{gomann_agrarrelevante_2015} and made available in Table 2.

\begin{longtable}[]{@{}
		>{\raggedright\arraybackslash}p{(\linewidth - 8\tabcolsep) * \real{0.2320}}
		>{\raggedright\arraybackslash}p{(\linewidth - 8\tabcolsep) * \real{0.1800}}
		>{\raggedright\arraybackslash}p{(\linewidth - 8\tabcolsep) * \real{0.2000}}
		>{\raggedright\arraybackslash}p{(\linewidth - 8\tabcolsep) * \real{0.1480}}
		>{\raggedright\arraybackslash}p{(\linewidth - 8\tabcolsep) * \real{0.2400}}@{}}
	\caption{Weather variables (indicators) provided by the German
		Weather Service (DWD), selectively used as features in the machine learning model and for calculation of extreme weather indicators.}
	\tabularnewline
	\toprule\noalign{}
	\begin{minipage}[b]{\linewidth}\raggedright
		\textbf{Indicator}
	\end{minipage} &
	\begin{minipage}[b]{\linewidth}\raggedright
		\textbf{Frequency Options}
	\end{minipage} &
	\begin{minipage}[b]{\linewidth}\raggedright
		\textbf{Variable Type}
	\end{minipage} &
	\begin{minipage}[b]{\linewidth}\raggedright
		\textbf{Units}
	\end{minipage} &
	\begin{minipage}[b]{\linewidth}\raggedright
		\textbf{Notes}
	\end{minipage} \\
	\midrule\noalign{}
	\endfirsthead
	
	\toprule\noalign{}
	\begin{minipage}[b]{\linewidth}\raggedright
		\textbf{Indicator}
	\end{minipage} &
	\begin{minipage}[b]{\linewidth}\raggedright
		\textbf{Frequency Options}
	\end{minipage} &
	\begin{minipage}[b]{\linewidth}\raggedright
		\textbf{Variable Type}
	\end{minipage} &
	\begin{minipage}[b]{\linewidth}\raggedright
		\textbf{Units}
	\end{minipage} &
	\begin{minipage}[b]{\linewidth}\raggedright
		\textbf{Notes}
	\end{minipage} \\
	\midrule\noalign{}
	\endhead
	
	\bottomrule\noalign{}
	\endlastfoot
	
	Mean Temperature & Daily, Weekly, Monthly & Temperature & °C & Core input for PET and temperature-related metrics \\
	Max Temperature & Daily, Weekly, Monthly & Temperature & °C & Used in Wechselfrost and heat indicators \\
	Min Temperature & Daily, Weekly, Monthly & Temperature & °C & Used in frost and Wechselfrost indicators \\
	Precipitation & Daily, Weekly, Monthly & Precipitation & mm/day & Used in SPI, SPEI, Gewitter, and general analysis \\
	Wind Speed & Daily, Weekly, Monthly & Wind Speed & m/s & Used in Gewitter and PET \\
	Solar Radiation & Daily, Weekly, Monthly & Radiation & W/m² or MJ/m²/day & Used in PET calculations \\
	Relative Humidity & Daily, Weekly, Monthly & Humidity & \% & Used in PET \\
	PET (Penman–Monteith) & Daily, Weekly, Monthly & Derived (from multiple variables) & mm/day & Used in SPEI \\
\end{longtable}

\pagebreak

\begin{longtable}[]{@{}
		>{\raggedright\arraybackslash}p{(\linewidth - 6\tabcolsep) * \real{0.2980}}
		>{\raggedright\arraybackslash}p{(\linewidth - 6\tabcolsep) * \real{0.1744}}
		>{\raggedright\arraybackslash}p{(\linewidth - 6\tabcolsep) * \real{0.2598}}
		>{\raggedright\arraybackslash}p{(\linewidth - 6\tabcolsep) * \real{0.2678}}@{}}
	\caption{Extreme weather indicators derived from daily weather
		variables. These were used selectively as features in the machine learning models.}\tabularnewline
	\toprule\noalign{}
	\begin{minipage}[b]{\linewidth}\raggedright
		\textbf{Indicator}
	\end{minipage} &
	\begin{minipage}[b]{\linewidth}\raggedright
		\textbf{Frequency Options}
	\end{minipage} &
	\begin{minipage}[b]{\linewidth}\raggedright
		\textbf{Predictor Type}
	\end{minipage} &
	\begin{minipage}[b]{\linewidth}\raggedright
		\textbf{Threshold Type}
	\end{minipage} \\
	\midrule\noalign{}
	\endfirsthead
	
	\toprule\noalign{}
	\begin{minipage}[b]{\linewidth}\raggedright
		\textbf{Indicator}
	\end{minipage} &
	\begin{minipage}[b]{\linewidth}\raggedright
		\textbf{Frequency Options}
	\end{minipage} &
	\begin{minipage}[b]{\linewidth}\raggedright
		\textbf{Predictor Type}
	\end{minipage} &
	\begin{minipage}[b]{\linewidth}\raggedright
		\textbf{Threshold Type}
	\end{minipage} \\
	\midrule\noalign{}
	\endhead
	
	\bottomrule\noalign{}
	\endlastfoot
	
	Frost Days & Weekly, Monthly & Temperature & Absolute \\
	Heat Days (\textgreater{} Threshold) & Weekly, Monthly & Temperature & Absolute \\
	Percentile-based Hot Days & Weekly, Monthly & Temperature & Percentile \\
	SPEI (Standardized Precip–Evap Index) & Monthly, Quarterly & Composite (Water Balance) & Distribution \\
	SPI (Standardized Precip Index) & Monthly, Quarterly & Precipitation & Distribution \\
	High Radiation Days & Monthly & Radiation & Absolute (e.g., \textgreater 2500 J/cm²) \\
	Days in 99th Percentile Precipitation & Monthly & Precipitation & Percentile (99th) \\
	Days in 99th Percentile Wind Speed & Monthly & Wind & Percentile (99th) \\
	Frost change (Wechselfrost) Days & Monthly & Temperature & Custom \\
	Thunder Storm (Gewitter) Days & Monthly & Precipitation + Wind & Custom \\
	
\end{longtable}

\subsection{Yield Data Preprocessing}\label{yield-data-preprocessing}

Data was pre-processed to suit the machine learning model setup. The
average yield was calculated in each region by removing the
2\textsuperscript{nd} order polynomial average German yield trend from
the observed NUTS-3 yield, and adding the maximum polynomial yield.
Polynomial trend was calculated based on maximum average national yield
observed in the study region \citep{lobell_climate_2011, volker_long-term_2022}. The variation in
percent of the ratio between actual and potential yield across years was
calculate to indicate how regions approach their potential yield. The
absolute yield gap is the difference between the maximum detrended yield
over 1979--2022 for each NUTS-3 region and the actual annual detrended
yield with a typical range between -30 and +30 percent. The yield gap
was the percentage difference between the mean detrended yield over
1978--2028 and the actual annual detrended yield for each NUTS-3 region
with a range between zero and a few tonnes per hectares. Yield anomaly
was calculated based on NUTS-3 specific rolling mean ten year historical
yield (lagged by 2 years) with a range of +/- 10 percent around zero
percent. An overview on key variables, variability and extreme values of
yield in German NUTS-3 regions is provided in Figure 1.

\begin{figure}
\centering
\includegraphics[width=5.96in,height=4.852in]{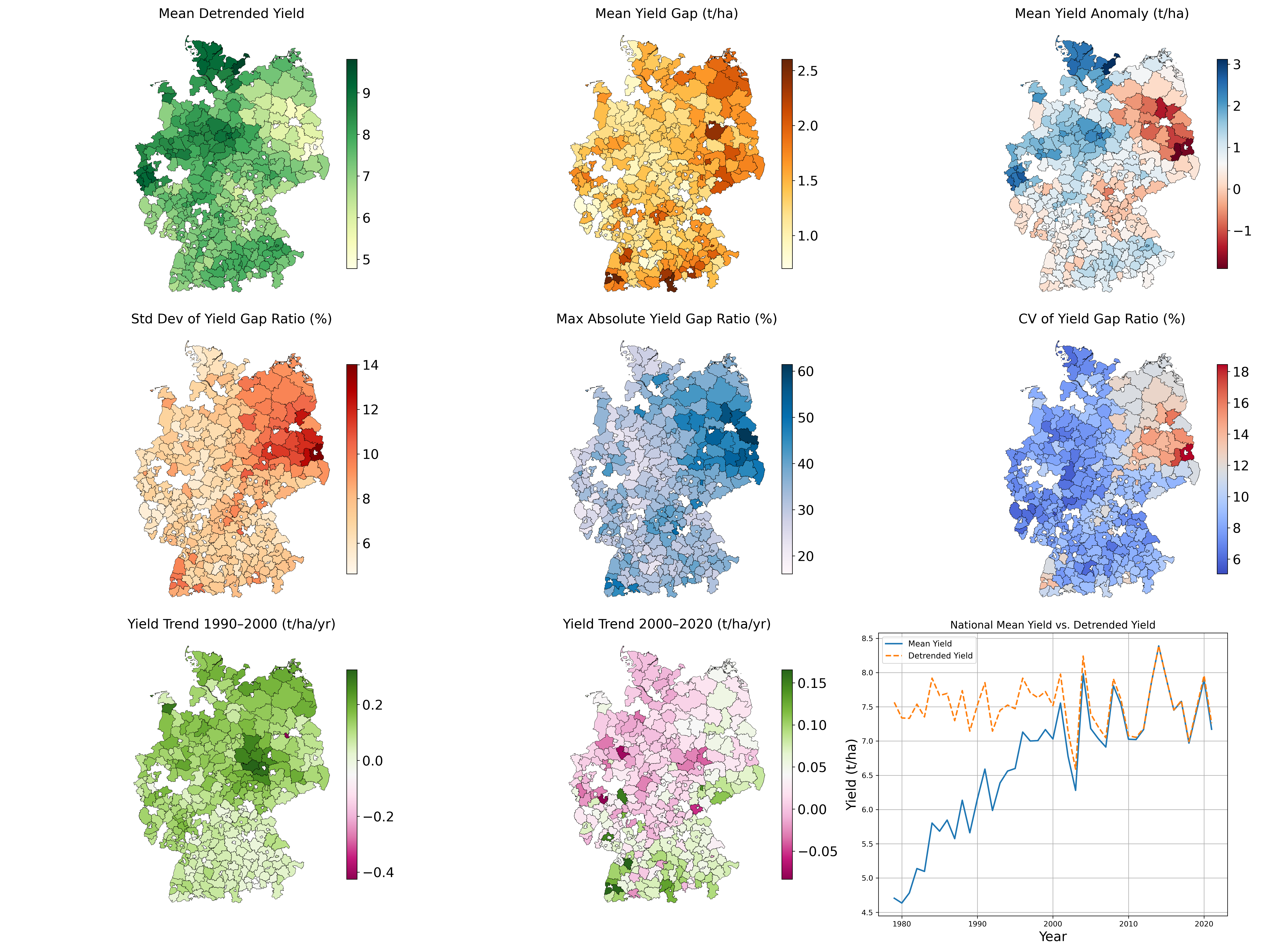}
\caption{Winter wheat yield key statistics for Germany. Yield
gap, anomaly, gap ratio, maximum absolute gap ratio, coefficient of
variation of yield gap ratio are based on detrended yield. Yield trends
for two period were calculated for observed yield. Finally, observed
(reported) yield and detrended yield were reported.}
\end{figure}

\subsection{Machine Learning Model
Setup}\label{machine-learning-model-setup}

The four machine learning models XGBoost (XGB), Random Forest (RF), Long
Short Term Memory Network (LSTM) and Temporal Convolutional Network
(TCN) were used. The models group into ensemble tree based prediction
(XGB and RF) on two dimensional inputs and explicit time series
regression models (LSTM and TCN). Tree-based and deep learning models
are consistently used in crop yield and anomaly prediction \citep{lischeid_machine_2022,cao_mapping_2025, jabed_crop_2024}, while direct comparisons are rare and
difficult to assess in a consistent manner \citep{patil_comparative_2025}. All
models were provided different feature setups to inform yield
prediction. Feature sets ranged from one (reference run for tree-based
models) to many features. Tree-based models were set up with pair-wise
congruent features to allow direct comparison of feature importance, and
deep learning models were set up with pair-wise congruent temporal
feature sets as well.

The reference run was made with a single predictor by region i.e. NUTS-3
mean regional yield. Further set ups include for example only Standard
Precipitation Index for the first nine months, each monthly value
representing a single feature, or mean air temperature for the first
nine months of each year, or stacking several of these at monthly or
weekly resolution together for a large multi-feature ML model. Time
series prediction algorithms were provided with time-constant features
and time series features e.g. 9 months of SPI only, or combining several
of the temporal variables. Here, always the full time series was used as
one temporally changing feature.

Analogues to other studies to predict yield metrics, all four algorithms
were set up to predict absolute yield \citep{tamayo-vera_advanced_2025, paudel_machine_2022}, yield gap \citep{cao_mapping_2025, mousavi_spatial_2024}, yield gap ratio \citep{brill_exploring_2024} and total
yield anomaly \citep{lischeid_machine_2022, jabed_crop_2024}. Model performance was quantified
with the root mean square error and the coefficient of determination:

\(R^{2} = 1 - \frac{{SS}_{model}}{{SS}_{naive\ mean}}\) (1)

Observational and input data was split into four data sets. Validation
Data were chose as two years of rather high (year: 2004) and rather low
(year: 2018) nation-wide average crop yield. Test Data was 10 percent
randomly sampled amongst NUTS-3 regions and remaining years (2000-2022).
ML Models than split internally the remaining data into training and
testing data. Here, testing data will be similarly sampled as Test Data.

Hyperparameter optimization was systematically applied to all four
models to ensure robust performance \citep{zoller_benchmark_2019}. For Random
Forest and XGBoost, grid search was conducted using GridSearchCV with
time-series-aware cross-validation. For deep learning models (LSTM and
TCN), KerasTuner \citep{omalley2020kerastuner} was employed to explore
architectures and training configurations, including the number of
layers, units, dropout rates, and learning rates. Each model was
optimized independently to minimize root mean squared error (RMSE) on
the validation set. SHAP (SHapley Additive exPlanations) were calculated
for XGBoost and Random Forest models to assess feature importance based
on the feature's contribution to the prediction \citep{lundberg_local_2020}.

\section{Results}\label{results}

All model results provided positive training and testing results,
representing convergence. The following sections report the evaluation
results on Test Data and Validation Data. Based on these results, the
SHAP analysis results are presented with valid and invalid feature
importance ratings.

\subsection{Model performance}\label{model-performance}

Model performance for all Test Data were associated with a positive R2
for the four ML models. This indicates a positive and generalized model
with regard to availability of information within the region at the time
of prediction. Evaluation with the Validation Data provided contrasting
performance. All ML Models included scenarios were R2 Validation was
below 0 while R2 Test was above zero. This is an indicator that the mean
of all simulations in non-regular yield years is worse than the
prediction given no spatial yield observations. However, there are also
many models for ensemble tree based as well as deep learning models that
resulted in positive R2 Validation. Another result is that correlation
between R2 Test and R2 Validation is similar and positive for both
ensemble tree based models. Albeit, it must be noted that high R2 Test
does not necessarily result in a high R2 Validation but also can result
in a negative R2 Validation. Correlation between R2 Test and R2
Validation is dissimilar for the categories of ensemble tree based
models (positive slope) and deep learning models (negative slope).
Across all models, best model performances were achieved independent of
the ML algorithm category. Reference performance noted by black asterisk
(Random Forest) and circle (XGBoost) were relatively high for R2 Test
and R2 Validation, and only few models outperformed these simple models.
Spatial variation is high, anomalies are spatially correlated,
prediction of temporal dynamics is more challenging compared prediction
of spatial variance and many models perform well with for predicting
spatial patterns but underperform in prediction of temporal dynamics
i.e. anomalies. The following section examines the results of feature
importance in the light of ML performance or underperformance.

\begin{figure}
\centering
\includegraphics[width=6.3in,height=5.87407in]{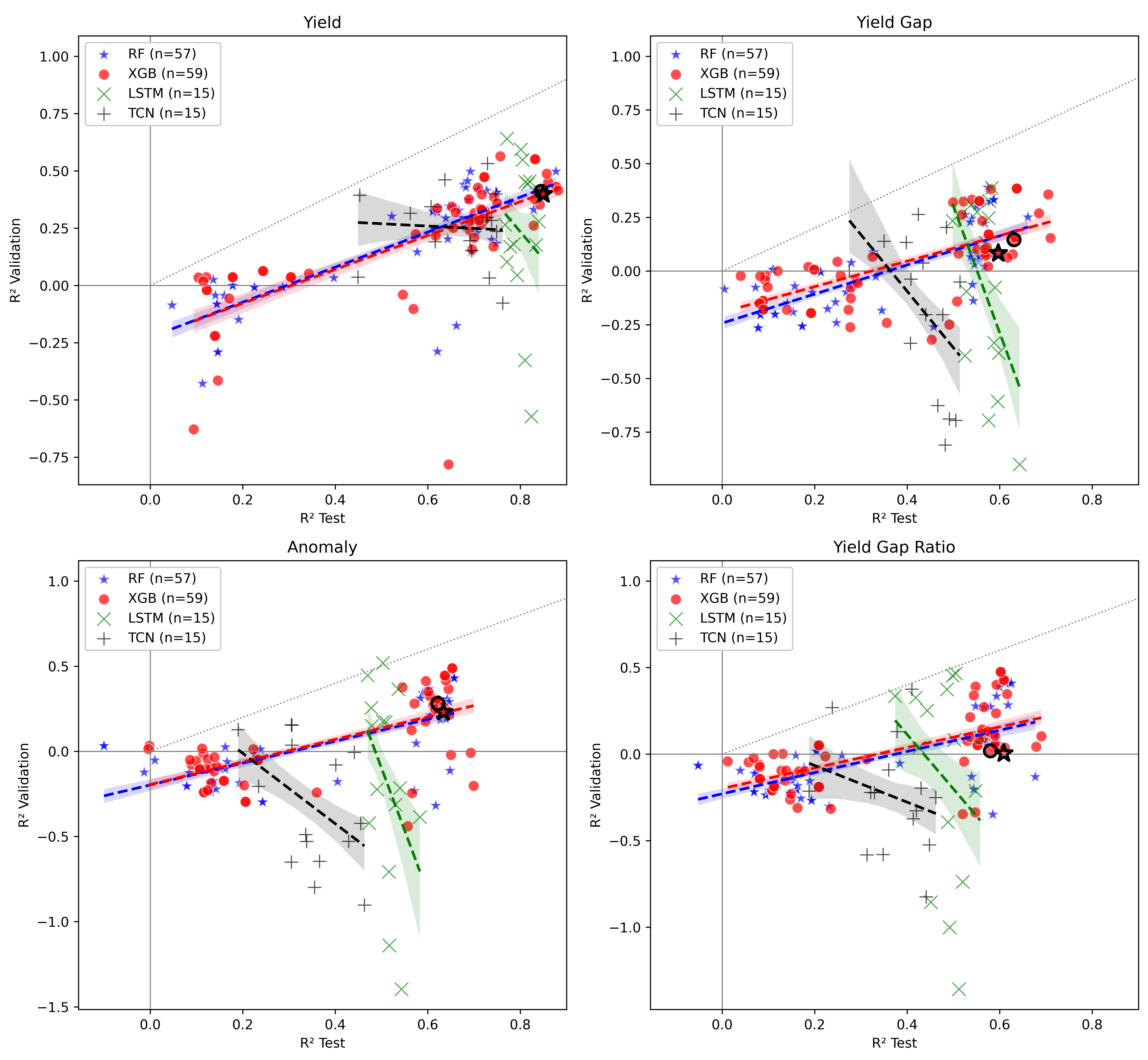}
\caption{Model Performance for various regression models of
XGBoost (XGB), Random Forest (RF), Long Short Term Memory Network (LSTM)
and Temporal Convolutional Network (TCN) measured by coefficient of
determination: R2 Test measures testing predictions given knowledge of
NUTS-3 yield in the year of prediction, while R2 Validation measures
prediction accuracy focusing on temporal dynamics without knowledge of
NUTS-3 yield in any region at the two years in question. Black asterisk
and circle are reference model performance.}
\end{figure}

\subsection{Feature importance in the light of ML performance or
underperformance}\label{feature-importance-in-the-light-of-ml-performance-or-underperformance}

SHapley Additive exPlanations (SHAP) values were calculated for ML
models independent of their R2 value and with paired feature setups.
Feature importance in terms of SHAP indicate a strong correlation
between the two tree-based ML algorithms XGBoost and Random Forest
(Figure 2). Hence, the capacity to explain degrading model behavior or
even underperforming model performance was attributed to many features
for a multitude of ML models. The results demonstrate a tendency for
degrading and underperforming models to provide rather high SHAP values
compared to those of effective model runs. These results depend on the
model feature setups, indicating that underperforming and degrading
models exhibit rather less features than effective models. Effective
model's feature importance is associated with few and many features,
though the majority of effective model's feature importance was
associated with small SHAP values.

\begin{figure}
\centering
\includegraphics[width=6.3in,height=5.61608in]{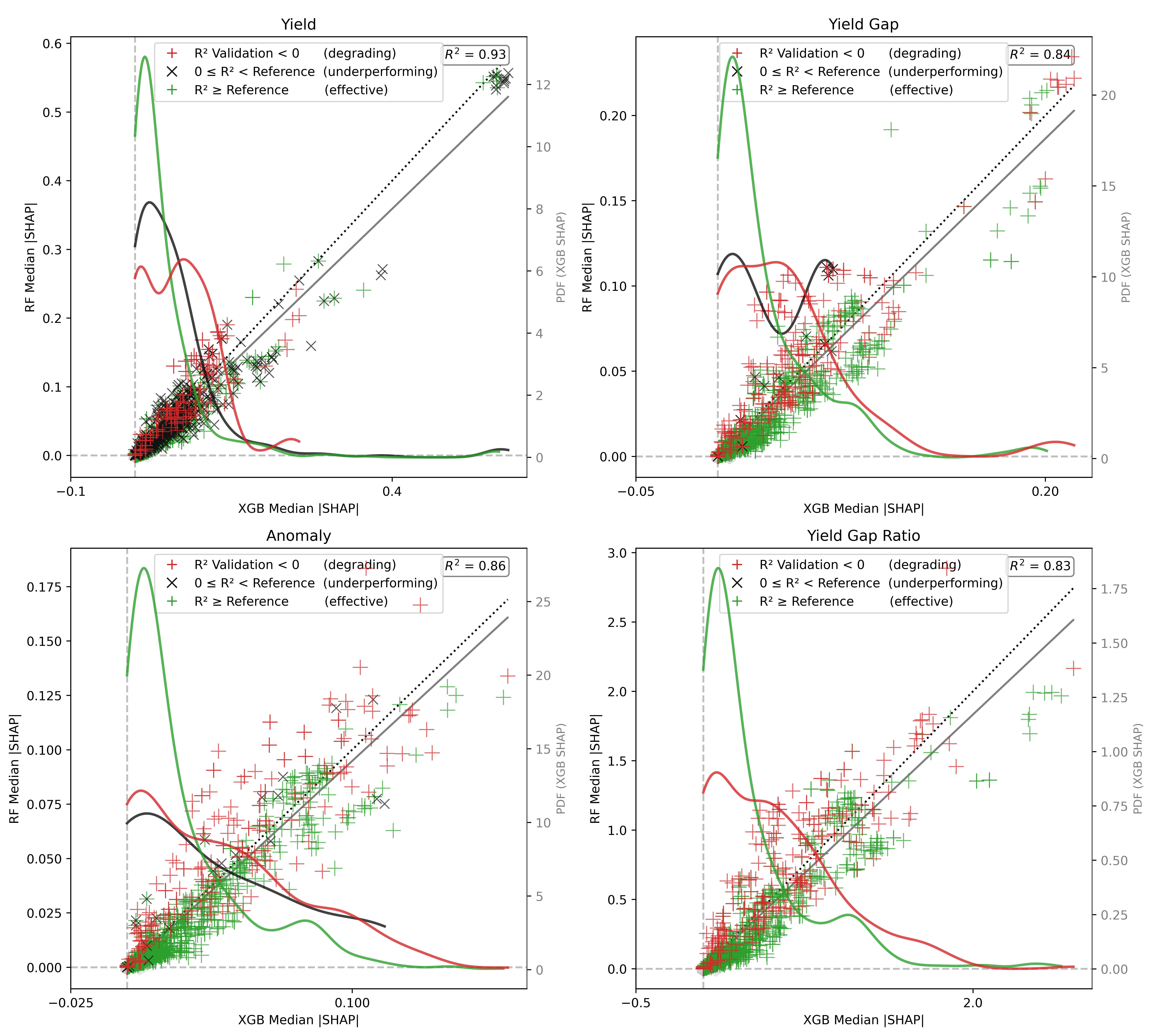}
\caption{Feature importance quantified by SHAP values for
effective, degrading and underperforming ML model. Predictions were made
for yield, yield gap, yield gap ratio and yield anomaly using various
Random Forest and XGBoost models.}
\end{figure}

\section{Discussion}\label{discussion}

\subsection{Model Generalization and
Validation}\label{model-generalization-and-validation}

Across all models, test set performance showed generally positive R²
values, indicating that models successfully captured spatial yield
variance when evaluated on random samples of space-time data. However,
the models' performance on the independent validation years (2004 and
2018) exposed frequent drops in R², sometimes to negative values. This
inconsistency underlines a key issue: High performance on test sets does
not guarantee model generalization to unseen conditions. Particularly,
temporal extrapolation appears far more challenging than spatial
interpolation. Spatial and temporal context is critical and should guide
the validation process in crop and machine learning models
\citep{reichstein_deep_2019}.

\subsection{Tree-based versus deep learning
models}\label{tree-based-versus-deep-learning-models}

Interestingly, the tree-based models (RF and XGBoost) demonstrated
better stability than deep learning models (LSTM, TCN) noted by the
positive correlation of space-and-time-validation. The negative
correlation between R² Test and R² Validation for deep learning models
may be attributed to their increased flexibility towards temporal
behavior, both methods achieving similar or at times better performance
in validation years but consistently worse performance in spatial
testing. This is a notable difference, potentially positioning deep
learning models to better simulate dynamics than tree-based models
\citep{reichstein_deep_2019,jabed_crop_2024}.

\subsection{Interpreting Feature Importance in Light of Model
Performance}\label{interpreting-feature-importance-in-light-of-model-performance}

The SHAP analysis revealed an important and under-discussed phenomenon:
Even models with low or negative validation performance produced
seemingly interpretable feature importance results. Particularly,
underperforming models often assigned disproportionately high SHAP
values to a narrow set of features. This highlights a critical issue:
Interpretation of ML models is only valid if the models themselves are
demonstrably generalizable. This is too often overlooked although
generally perceived and discussed \citep{letoffe_shap_2024,salih_perspective_2025}. Without
robust validation, such interpretations risk overfitting and false
attribution of explanatory power to input variables. Given the high R2
Test and low R2 validation for some models in anomaly and yield gap
ratio, this is particularly important finding for interpretation of
related studies \citep{lischeid_machine_2022, cao_mapping_2025,lundberg_local_2020}.

The fact that effective models often displayed lower average SHAP values
per feature dispersed across many inputs suggests that stable models
better capture the nuanced interactions among variables. In contrast,
unstable models may rely excessively on spurious correlations, leading
to inflated SHAP values for a small subset of features
\citep{letoffe_shap_2024, salih_perspective_2025}.

\section{Conclusions}\label{conclusions}

This study provides a detailed evaluation of machine learning (ML)
models for predicting agricultural yield metrics and anomalies using
both ensemble tree-based and deep learning approaches. Our findings
highlight the risks of overfitting and misinterpretation, particularly
when model generalization is not rigorously tested.

A central insight is that feature importance interpretations from ML
models can be misleading if the underlying models are not demonstrably
generalizable. Models that perform well on test data but poorly on
temporally independent validation data often produce inflated or
inconsistent importance scores. This raises concerns about drawing
agronomic conclusions from such outputs, particularly in studies where
no independent validation is performed.

These results underscore the importance of rigorous, context-specific
validation protocols---especially those that test temporal extrapolation
under conditions of anomalous yields. Relying exclusively on random data
splits or cross-validation may provide overly optimistic assessments of
model skill and explanatory value.

To support robust and interpretable ML applications in agriculture,
following recommendations can be drawn: (i) incorporating domain
knowledge into model design and feature selection, (ii) transparently
documenting validation strategies, including the use of temporally
independent data and finally (iii) exercising caution when interpreting
feature importance, particularly in the absence of proven model
generalization.

Furthermore, the observed challenge in capturing temporal dynamics
suggests that standalone ML models may have structural limitations.
Hybrid approaches that combine process-based models with data-driven
learning may offer a promising path toward more resilient and
interpretable yield prediction systems.

Ultimately, progress in ML-based yield modeling hinges not just on
achieving high predictive performance, but on building credible,
transparent, and reproducible frameworks that support decision-making in
agricultural systems under changing environmental conditions.

\bibliographystyle{agsm}
\bibliography{ML}

@article{lischeid_machine_2022,
	title = {Machine learning in crop yield modelling: {A} powerful tool, but no surrogate for science},
	volume = {312},
	issn = {01681923},
	shorttitle = {Machine learning in crop yield modelling},
	url = {https://linkinghub.elsevier.com/retrieve/pii/S0168192321003841},
	doi = {10.1016/j.agrformet.2021.108698},
	language = {en},
	urldate = {2025-11-18},
	journal = {Agricultural and Forest Meteorology},
	author = {Lischeid, Gunnar and Webber, Heidi and Sommer, Michael and Nendel, Claas and Ewert, Frank},
	month = jan,
	year = {2022},
	pages = {108698},
}

@article{lobell_climate_2011,
	title = {Climate {Trends} and {Global} {Crop} {Production} {Since} 1980},
	volume = {333},
	issn = {0036-8075, 1095-9203},
	url = {https://www.science.org/doi/10.1126/science.1204531},
	doi = {10.1126/science.1204531},
	language = {en},
	number = {6042},
	urldate = {2025-11-18},
	journal = {Science},
	author = {Lobell, David B. and Schlenker, Wolfram and Costa-Roberts, Justin},
	month = jul,
	year = {2011},
	pages = {616--620},
}

@article{cao_mapping_2025,
	title = {Mapping global yields of four major crops at 5-minute resolution from 1982 to 2015 using multi-source data and machine learning},
	volume = {12},
	issn = {2052-4463},
	url = {https://www.nature.com/articles/s41597-025-04650-4},
	doi = {10.1038/s41597-025-04650-4},
	language = {en},
	number = {1},
	urldate = {2025-11-18},
	journal = {Scientific Data},
	author = {Cao, Juan and Zhang, Zhao and Luo, Xiangzhong and Luo, Yuchuan and Xu, Jialu and Xie, Jun and Han, Jichong and Tao, Fulu},
	month = feb,
	year = {2025},
	pages = {357},
}

@article{tamayo-vera_advanced_2025,
	title = {Advanced machine learning for regional potato yield prediction: analysis of essential drivers},
	volume = {3},
	issn = {2731-9202},
	shorttitle = {Advanced machine learning for regional potato yield prediction},
	url = {https://www.nature.com/articles/s44264-025-00052-6},
	doi = {10.1038/s44264-025-00052-6},
	language = {en},
	number = {1},
	urldate = {2025-11-18},
	journal = {npj Sustainable Agriculture},
	author = {Tamayo-Vera, Dania and Mesbah, Morteza and Zhang, Yinsuo and Wang, Xiuquan},
	month = mar,
	year = {2025},
	pages = {12},
}

@article{reichstein_deep_2019,
	title = {Deep learning and process understanding for data-driven {Earth} system science},
	volume = {566},
	issn = {0028-0836, 1476-4687},
	url = {https://www.nature.com/articles/s41586-019-0912-1},
	doi = {10.1038/s41586-019-0912-1},
	language = {en},
	number = {7743},
	urldate = {2025-11-11},
	journal = {Nature},
	author = {Reichstein, Markus and Camps-Valls, Gustau and Stevens, Bjorn and Jung, Martin and Denzler, Joachim and Carvalhais, Nuno and {Prabhat}},
	month = feb,
	year = {2019},
	pages = {195--204},
}

@article{karpatne_theory-guided_2017,
	title = {Theory-{Guided} {Data} {Science}: {A} {New} {Paradigm} for {Scientific} {Discovery} from {Data}},
	volume = {29},
	copyright = {https://ieeexplore.ieee.org/Xplorehelp/downloads/license-information/IEEE.html},
	issn = {1041-4347},
	shorttitle = {Theory-{Guided} {Data} {Science}},
	url = {http://ieeexplore.ieee.org/document/7959606/},
	doi = {10.1109/TKDE.2017.2720168},
	number = {10},
	urldate = {2025-11-11},
	journal = {IEEE Transactions on Knowledge and Data Engineering},
	author = {Karpatne, Anuj and Atluri, Gowtham and Faghmous, James H. and Steinbach, Michael and Banerjee, Arindam and Ganguly, Auroop and Shekhar, Shashi and Samatova, Nagiza and Kumar, Vipin},
	month = oct,
	year = {2017},
	pages = {2318--2331},
}

@article{baatz_reanalysis_2021,
	title = {Reanalysis in {Earth} {System} {Science}: {Toward} {Terrestrial} {Ecosystem} {Reanalysis}},
	volume = {59},
	issn = {8755-1209, 1944-9208},
	shorttitle = {Reanalysis in {Earth} {System} {Science}},
	url = {https://agupubs.onlinelibrary.wiley.com/doi/10.1029/2020RG000715},
	doi = {10.1029/2020RG000715},
	language = {en},
	number = {3},
	urldate = {2025-11-18},
	journal = {Reviews of Geophysics},
	author = {Baatz, R. and Hendricks Franssen, H. J. and Euskirchen, E. and Sihi, D. and Dietze, M. and Ciavatta, S. and Fennel, K. and Beck, H. and De Lannoy, G. and Pauwels, V. R. N. and Raiho, A. and Montzka, C. and Williams, M. and Mishra, U. and Poppe, C. and Zacharias, S. and Lausch, A. and Samaniego, L. and Van Looy, K. and Bogena, H. and Adamescu, M. and Mirtl, M. and Fox, A. and Goergen, K. and Naz, B. S. and Zeng, Y. and Vereecken, H.},
	month = sep,
	year = {2021},
	pages = {e2020RG000715},
}

@article{baatz_evaluation_2017,
	title = {Evaluation of a cosmic-ray neutron sensor network for improved land surface model prediction},
	volume = {21},
	copyright = {https://creativecommons.org/licenses/by/3.0/},
	issn = {1607-7938},
	url = {https://hess.copernicus.org/articles/21/2509/2017/},
	doi = {10.5194/hess-21-2509-2017},
	language = {en},
	number = {5},
	urldate = {2025-11-18},
	journal = {Hydrology and Earth System Sciences},
	author = {Baatz, Roland and Hendricks Franssen, Harrie-Jan and Han, Xujun and Hoar, Tim and Bogena, Heye Reemt and Vereecken, Harry},
	month = may,
	year = {2017},
	pages = {2509--2530},
}

@article{vrugt_treatment_2008,
	title = {Treatment of input uncertainty in hydrologic modeling: {Doing} hydrology backward with {Markov} chain {Monte} {Carlo} simulation},
	volume = {44},
	copyright = {http://onlinelibrary.wiley.com/termsAndConditions\#vor},
	issn = {0043-1397, 1944-7973},
	shorttitle = {Treatment of input uncertainty in hydrologic modeling},
	url = {https://agupubs.onlinelibrary.wiley.com/doi/10.1029/2007WR006720},
	doi = {10.1029/2007WR006720},
	language = {en},
	number = {12},
	urldate = {2025-11-18},
	journal = {Water Resources Research},
	author = {Vrugt, Jasper A. and Ter Braak, Cajo J. F. and Clark, Martyn P. and Hyman, James M. and Robinson, Bruce A.},
	month = dec,
	year = {2008},
	pages = {2007WR006720},
}

@article{zoller_benchmark_2019,
	title = {Benchmark and {Survey} of {Automated} {Machine} {Learning} {Frameworks}},
	copyright = {arXiv.org perpetual, non-exclusive license},
	url = {https://arxiv.org/abs/1904.12054},
	doi = {10.48550/ARXIV.1904.12054},
	urldate = {2025-11-18},
	author = {Zöller, Marc-André and Huber, Marco F.},
	year = {2019},
	note = {Publisher: arXiv
Version Number: 5},
}

@misc{letoffe_shap_2024,
	title = {{SHAP} scores fail pervasively even when {Lipschitz} succeeds},
	copyright = {Creative Commons Attribution 4.0 International},
	url = {https://arxiv.org/abs/2412.13866},
	doi = {10.48550/ARXIV.2412.13866},
	urldate = {2025-11-18},
	publisher = {arXiv},
	author = {Letoffe, Olivier and Huang, Xuanxiang and Marques-Silva, Joao},
	year = {2024},
	note = {Version Number: 1},
}

@article{jabed_crop_2024,
	title = {Crop yield prediction in agriculture: {A} comprehensive review of machine learning and deep learning approaches, with insights for future research and sustainability},
	volume = {10},
	issn = {24058440},
	shorttitle = {Crop yield prediction in agriculture},
	url = {https://linkinghub.elsevier.com/retrieve/pii/S2405844024168673},
	doi = {10.1016/j.heliyon.2024.e40836},
	language = {en},
	number = {24},
	urldate = {2025-11-18},
	journal = {Heliyon},
	author = {Jabed, Md. Abu and Azmi Murad, Masrah Azrifah},
	month = dec,
	year = {2024},
	pages = {e40836},
}

@article{webber_no_2020,
	title = {No perfect storm for crop yield failure in {Germany}},
	volume = {15},
	issn = {1748-9326},
	url = {https://iopscience.iop.org/article/10.1088/1748-9326/aba2a4},
	doi = {10.1088/1748-9326/aba2a4},
	number = {10},
	urldate = {2025-11-18},
	journal = {Environmental Research Letters},
	author = {Webber, Heidi and Lischeid, Gunnar and Sommer, Michael and Finger, Robert and Nendel, Claas and Gaiser, Thomas and Ewert, Frank},
	month = oct,
	year = {2020},
	pages = {104012},
}

@article{seidel_towards_2018,
	title = {Towards improved calibration of crop models – {Where} are we now and where should we go?},
	volume = {94},
	issn = {11610301},
	url = {https://linkinghub.elsevier.com/retrieve/pii/S1161030118300066},
	doi = {10.1016/j.eja.2018.01.006},
	language = {en},
	urldate = {2025-11-18},
	journal = {European Journal of Agronomy},
	author = {Seidel, S.J. and Palosuo, T. and Thorburn, P. and Wallach, D.},
	month = mar,
	year = {2018},
	pages = {25--35},
}

@article{brill_exploring_2024,
	title = {Exploring drought hazard, vulnerability, and related impacts on agriculture in {Brandenburg}},
	volume = {24},
	copyright = {https://creativecommons.org/licenses/by/4.0/},
	issn = {1684-9981},
	url = {https://nhess.copernicus.org/articles/24/4237/2024/},
	doi = {10.5194/nhess-24-4237-2024},
	language = {en},
	number = {12},
	urldate = {2025-11-18},
	journal = {Natural Hazards and Earth System Sciences},
	author = {Brill, Fabio and Alencar, Pedro Henrique Lima and Zhang, Huihui and Boeing, Friedrich and Hüttel, Silke and Lakes, Tobia},
	month = nov,
	year = {2024},
	pages = {4237--4265},
}

@misc{volker_long-term_2022,
	title = {Long-term crop yields, cultivation areas and total arable land in {Germany} at {NUTS} 3 level},
	copyright = {Creative Commons Attribution 4.0 International},
	doi = {10.4228/ZALF-MFW5-XG49},
	urldate = {2025-11-18},
	publisher = {Leibniz Centre for Agricultural Landscape Research (ZALF)},
	author = {Völker, Lidia and Ahrends, Ellen H. and Sommer, Michael and {Ellen H. Ahrends}},
	collaborator = {{Lidia Völker} and {Michael Sommer}},
	month = jun,
	year = {2022},
}

@article{blickensdorfer_mapping_2022,
	title = {Mapping of crop types and crop sequences with combined time series of {Sentinel}-1, {Sentinel}-2 and {Landsat} 8 data for {Germany}},
	volume = {269},
	issn = {00344257},
	url = {https://linkinghub.elsevier.com/retrieve/pii/S0034425721005514},
	doi = {10.1016/j.rse.2021.112831},
	language = {en},
	urldate = {2025-11-18},
	journal = {Remote Sensing of Environment},
	author = {Blickensdörfer, Lukas and Schwieder, Marcel and Pflugmacher, Dirk and Nendel, Claas and Erasmi, Stefan and Hostert, Patrick},
	month = feb,
	year = {2022},
	pages = {112831},
}

@book{gomann_agrarrelevante_2015,
	address = {DE},
	title = {Agrarrelevante {Extremwetterlagen} und {Möglichkeiten} von {Risikomanagementsystemen} : {Studie} im {Auftrag}},
	shorttitle = {Agrarrelevante {Extremwetterlagen} und {Möglichkeiten} von {Risikomanagementsystemen}},
	url = {https://doi.org/10.3220/REP1434012425000},
	language = {ger},
	urldate = {2025-11-18},
	publisher = {Johann Heinrich von Thünen-Institut},
	author = {Gömann, Horst and Bender, Andrea and Bolte, Andreas},
	year = {2015},
}

@article{patil_comparative_2025,
	title = {Comparative {Analysis} of {Machine} {Learning} {Models} for {Crop} {Yield} {Prediction} {Across} {Multiple} {Crop} {Types}},
	volume = {6},
	issn = {2661-8907},
	url = {https://link.springer.com/10.1007/s42979-024-03602-w},
	doi = {10.1007/s42979-024-03602-w},
	language = {en},
	number = {1},
	urldate = {2025-11-18},
	journal = {SN Computer Science},
	author = {Patil, Yashraj and Ramachandran, Harikrishnan and Sundararajan, Sridhevi and Srideviponmalar, P.},
	month = jan,
	year = {2025},
	pages = {64},
}

@article{paudel_machine_2022,
	title = {Machine learning for regional crop yield forecasting in {Europe}},
	volume = {276},
	issn = {03784290},
	url = {https://linkinghub.elsevier.com/retrieve/pii/S0378429021003233},
	doi = {10.1016/j.fcr.2021.108377},
	language = {en},
	urldate = {2025-11-18},
	journal = {Field Crops Research},
	author = {Paudel, Dilli and Boogaard, Hendrik and De Wit, Allard and Van Der Velde, Marijn and Claverie, Martin and Nisini, Luigi and Janssen, Sander and Osinga, Sjoukje and Athanasiadis, Ioannis N.},
	month = feb,
	year = {2022},
	pages = {108377},
}

@article{mousavi_spatial_2024,
	title = {Spatial prediction of winter wheat yield gap: agro-climatic model and machine learning approaches},
	volume = {14},
	issn = {1664-462X},
	shorttitle = {Spatial prediction of winter wheat yield gap},
	url = {https://www.frontiersin.org/articles/10.3389/fpls.2023.1309171/full},
	doi = {10.3389/fpls.2023.1309171},
	urldate = {2025-11-18},
	journal = {Frontiers in Plant Science},
	author = {Mousavi, Seyed Rohollah and Jahandideh Mahjenabadi, Vahid Alah and Khoshru, Bahman and Rezaei, Meisam},
	month = jan,
	year = {2024},
	pages = {1309171},
}

@misc{omalley2020kerastuner,
  author       = {O'Malley, Tom and Bursztein, Elie and Long, Hector and Chollet, François},
  title        = {Keras Tuner},
  year         = {2020},
  howpublished = {\url{https://keras.io/keras_tuner/}},
  note         = {Version as of 2020},
}

@article{lundberg_local_2020,
	title = {From local explanations to global understanding with explainable {AI} for trees},
	volume = {2},
	issn = {2522-5839},
	url = {https://www.nature.com/articles/s42256-019-0138-9},
	doi = {10.1038/s42256-019-0138-9},
	language = {en},
	number = {1},
	urldate = {2025-11-18},
	journal = {Nature Machine Intelligence},
	author = {Lundberg, Scott M. and Erion, Gabriel and Chen, Hugh and DeGrave, Alex and Prutkin, Jordan M. and Nair, Bala and Katz, Ronit and Himmelfarb, Jonathan and Bansal, Nisha and Lee, Su-In},
	month = jan,
	year = {2020},
	pages = {56--67},
}

@article{salih_perspective_2025,
	title = {A {Perspective} on {Explainable} {Artificial} {Intelligence} {Methods}: {SHAP} and {LIME}},
	volume = {7},
	issn = {2640-4567, 2640-4567},
	shorttitle = {A {Perspective} on {Explainable} {Artificial} {Intelligence} {Methods}},
	url = {https://advanced.onlinelibrary.wiley.com/doi/10.1002/aisy.202400304},
	doi = {10.1002/aisy.202400304},
	language = {en},
	number = {1},
	urldate = {2025-11-18},
	journal = {Advanced Intelligent Systems},
	author = {Salih, Ahmed M. and Raisi‐Estabragh, Zahra and Galazzo, Ilaria Boscolo and Radeva, Petia and Petersen, Steffen E. and Lekadir, Karim and Menegaz, Gloria},
	month = jan,
	year = {2025},
	pages = {2400304},
}

\end{document}